# Failure to Mix: Large language models struggle to answer according to desired probability distributions

Ivy Yuqian Yang[1], David Yu Zhang[1]
ivy.yang@biostate.ai, dave.zhang@biostate.ai
[1]Biostate AI, Houston, TX 77054

**Abstract:** Scientific idea generation and selection requires exploration following a target probability distribution. In contrast, current AI benchmarks have objectively correct answers, and training large language models (LLMs) via reinforcement learning against these benchmarks discourages probabilistic exploration. Here, we conducted systematic experiments requesting LLMs to produce outputs following simple probabilistic distributions, and found that all modern LLMs tested grossly fail to follow the distributions. For example, requesting a binary output of "1" 49% of the time produces an answer of "0" nearly 100% of the time. This step function-like behavior of near-exclusively generating the output with marginally highest probability even overrules even strong in-built LLM biases.

There is increasing interest and economic investment in artificial intelligence (AI) systems that autonomously conduct novel scientific research [1-4]. If achieved, such scientist AI systems could initiate a positive feedback loop in which AI-generated scientific breakthroughs facilitate and accelerate further breakthroughs. Scientific research is a combination of (1) idea generation/selection and (2) idea validation/falsification. Curated AI benchmarks such as Humanity Last Exam and BixBench [5-8] have objectively correct answers, and this has made large language models (LLMs) highly proficient at idea verification/falsification based on data. However, idea generation and selection requires probabilistic exploration of an infinite parameter space. There is no objectively correct answer for a single LLM response when exploring ideas, but we can judge ensembles of LLM responses for output variety and reasonableness, as a proxy for creativity and taste.

In the idea generation and selection phase, a strategy or policy is explicitly or implicitly constructed and followed, in order to allocate finite resources to an excess of potential ideas (Fig. 1a). As a simplistic model, the strategy is a probability distribution of distributing Resources to ideas with along a single dimensional of Risk/Reward. Note that Risk and Reward must be correlated for Pareto-optimal ideas; High Risk and Low Reward ideas are filtered out from consideration. Different nations, organizations, or principal investigators choose different target probability distributions. A competent autonomous AI scientist must thus also be able to direct idea generation and selection to follow a desired probability distribution. Here, we show that all current LLMs fail to do so.

## Results:

We begin our experimental evaluation using the simplest probability distribution possible, a binary outcome with a single parameter $p$ (Fig. 1b) corresponding the probability of responding with "1". For clarity, we also redundantly specify that "0" should be returned with probability $(1-p)$. We define the response rate $r$ as the observed rate of returning "1" out of N independent API calls of a LLM following the prompt in Fig. 1b. For $p = 0.45$, N=100, and Gemini 2.5 Pro, the observed response rate $r$ is 0.02.

We next systematically characterized $r$ for a variety of $p$ values ranging from 0 to 1, using N=100 API calls for each $p$ value (Fig. 1c). To our surprise, the response curve is essentially a step function with threshold at $p$=0.5 for Gemini 2.5 Pro. Based on this result, we initially suspected that the default LLM temperature was too low, and next set the temperature to be the maximum allowed by the Gemini 2.5 Pro API (Temperature = 2). This had essentially no effect; the $r$ vs. $p$ curve remains a step function (Fig. 1d). We next tested 7 other LLMs to ensure that this is not a quirk or artifact specific to the Gemini 2.5 Pro model. **Every single LLM tested displayed a step function behavior** (Fig. 1e).

**Multiple Iterative Decisions.** Next, we explored the possibility that LLMs may trend towards $r = p$ when each prompt asks for multiple responses instead of one. There are two possible ways that this could happen: (1) either later responses to the same prompt trend individually towards the desired probability distributions, or (2) collectively the responses in a single output stream trend towards the desired independently identically distributed probability distribution.

Fig. 2a shows the revised prompt and the range of responses when $p=0.48$ (N=100), and D=2 decisions per call. To quantitate the deviation of the observed response rate $r$ from the intended probably $p$ for different LLM responses, we define the metric step-similarity S via trapezoid rule for total area between the observed $r$ curve and the $r=p$ diagonal, multiplied by 4 so that a perfect step function exhibits S=1. See Supplementary Materials for mathematical definition of S. For D=2 (two responses per LLM call), the first response (j=1) maintains near perfect step-function behavior with S=0.958 (Fig. 2b). This result suggests that LLMs consistently follow a step function-like behavior on their first decision. However, the second response (j=2) follows an unexpected zigzag pattern with the probability of selecting "1" nearing 1 at $p = 0.49$, and then resetting back to near 0 just after $p > 0.5$ (Fig. 2c). This observation is consistent with the LLM observing their own first response and then trying to compensate for their own potential bias via their second selection, aiming for the desired ensemble distribution of "1" outputs. The behavior of 4 other LLMs tested also show this zigzag response, indicating generality of the "self-correction" behavior of current LLMs.

Further increasing the number of decisions to D=3 shows that j=1 and j=2 response curves look highly similar to analogous j=1 and j=2 curves for D=2 (Fig. 2d). Interestingly, the j=3 response curve is even more complex with 2 significant non-monotonic decreases of $r$ with increasing $p$, indicating continued LLM adjustment of later responses to try to meet correct ensemble response rates. Looking at the mean response rate across all D=2 or D=3 responses (Fig. 2e), we do see Mean($r$) does appear to approach the expected linear $r = p$ behavior, with Kimi K2 performing the best for D=2. At D=3, Gemini 2.5 Pro's Mean($r$) further approaches linear, with S=0.246 (compared to S=0.501 at D=2).

To test whether S for Mean($r$) would asymptotically approach 0 for larger values of D, we next tested D=10 using Google Gemini 2.5 Pro (Fig. 3ab). We find that the step-likeness S of Mean($r$) does decrease to S=0.138, and is suggestive of $r \approx p$ at even larger values of D. Unexpectedly, analysis of the 9th and 10th responses shows that these individual responses still show significantly higher S values. This suggests that obtaining a target probability distribution by requesting large number of responses and taking only the last response may not be cost effective as it may require a very large number of ignored responses.

Although Mean($r$) approaches the values of $p$ in bulk, also contains subtle statistical differences that render it a poor substitute for true independently identical distribution. At $p=0.15$ for N=1000 runs with D=10, we should expect that some of these 1000 runs will randomly generate a larger number of "1" responses (e.g. 3 or 4). However, all five LLM models tested show less variability in the total number of "1" outputs, usually clustering heavily around exactly a single "1" output out of 10 (Fig. 3c). An unusual exception is Qwen 3, which produced zero "1" outputs in 98.8% of cases. For $p=0.45$, we likewise see extremely high frequencies for specific number of total "1" outputs out of 10. This suggests that obtaining a target probability distribution by requesting large number of responses and selecting one answer after randomly shuffling them may also be ineffective.

**LLM Performance on a Two-Parameter Discrete Target Probability Distribution.** Given the strong and generalizable behavior of LLMs to the biased coin flip question (single parameter discrete target probability distribution), we next tested to see if the problem framing and target probability distribution is the exception rather than the rule of what LLMs struggle with. We revised the question into a multiple choice question with 3 potential responses (Fig. 4a), generating a 2 parameter discrete target probability distribution while also changing the framing and wording of the prompt. To simplify the experiment, we fixed the requested probability $p$ of producing "1" as output at 40%, and revised the requested probability $q$ of producing "2" as output between 0% and 60%.

This experiment generates 3 regions and 2 edge cases: When $q < 20\%$, "0" is the response with the single highest probability. When $40\% > q > 20\%$, "1" is the response with the single highest probability. When $q > 40\%$, "2" is the response with the single highest probability. At $q = 20\%$, "0" and "1" tie for most likely; at $q = 40\%$, "1" and "2" tie for most likely. If the LLMs were trained to always produce the single most likely response, then we would expect the response curves to continue to be step function-like around these critical threshold $q$ values.

Experimentally, we find that the results are significantly more complex and model-dependent (Fig. 4bcd). The response rate of the LLM producing "2" is closest to a step function, with the exception of the Kimi K2 model. For the "1" output, all 5 models respond with near $r = 1$ between $q = 20\%$ and 40%, but also with significant positive $r$ values for $q = 10\%$, unlike previous results in Fig. 1-3. The response rate of LLMs for the "0" output is highly varied in the $q < 20\%$ region: although all models generate $r \approx 1$ at $q = 0\%$ (when the correct frequency should be 60%), at $q = 10\%$, the response rate $r$ for "0" ranges from 0% to 100% for the 5 different models. Given these unexpected results, we further hypothesized that the different LLMs are based against the "0" token, or biased against the last listed option (or potentially both).

**Word Choice and Word Position Bias.** There are known and documented biases of LLMs for certain words and numbers [9-13]. For example, the numbers 37, 42, and 73 are over-represented when LLMs are requested to pick a number between 1 and 100. The ordering of the different output options may also bias the LLMs to answering in a certain way. To test these hypotheses, we next generated 11 pairs of words and asked the LLMs to respond with either the word shown in red (i.e. red word) or the blue word at $p = 50\%$ for each (Fig. 5). We systematically observed the response rate $r$ of the models to the red word, when the red word was offered first (Fig. 5b) vs. second (Fig. 5c). Averaging the response rate $r$ for the red word in both positions gives the overall model's bias to the red word, with 50% indicating completely unbiased, 100% indicating complete bias to the red word, and 0% indicate complete bias to the blue word.

We observed that for Gemini 2.5 Pro, the three word pairs with strongest statistically significant biases are "1" (over "0"), "sun" (over "moon"), and "human" (over "alien"). Average over the two columns, we find that the overall mean response rate for the red word is 75.5% when the red word is placed first, vs. 62.6% when the red word is placed second, indicating a mild bias to the first word position. However, even the strongest word and position bias ("sun" in first position) was unable to overcome even a 1% difference in the target $p$ value: at $p = 49\%$, the response rate $r$ for all red words in the first position were observed to be 0% (Fig. 5d).

Repeating the $p = 50\%$ experiments for four other LLMs (Fig. 5e) generated a summary diagram of the different model's "personalities" and biases (Fig. 5f). Fitting to it's name, Anthropic's Sonnet 3.5 also exhibits the single strongest bias towards "human" (vs. "alien") across all word pairs. Interestingly, Sonnet 4.5 also strongly biases towards "moon", in contrast to Gemini 2.5 Pro's strong bias toward "sun". Another interesting finding is that Qwen 3 much more strongly biases towards the first position, with "day" being the only word bias that supersedes the position bias. The most consistent bias across all models is that they all favor "1" over "0" (with the exception of Qwen 3).

**Application Problems With Mixed Strategy Solutions.** The prompts posed to the LLMs in previous experiments were all structured as biased coin flip or card draw "toy problems," so we next tested if providing more real-world context to problems would change the behavior of LLMs. The following prompt describes a real bioinformatics problem that is frequently encountered for RNA sequencing [14, 15]:

> *You are assisting with a bioinformatics task, specifically aligning NGS short reads to genes for RNA expression analysis. You are given a read R that aligns equally well to both genes A and B, because A and B are high homology. So far, 1000 reads have been uniquely mapped to gene A, and 1500 reads have been uniquely mapped to gene B. You must choose to map read R to either gene A or gene B as part of your task, and you wish the final results to maximally mirror the true expression levels of A and B. You MUST pick A or B for this independent task at the end of your response.*

The optimal response for the LLM is to probabilistically return gene A with $p=40\%$ and gene B with $(1-p)=60\%$ probability, and the $p$ value can be tuned via the relative numbers of uniquely mapped reads to genes A and B. Interestingly, all 5 LLMs challenged with the above problem were able correctly reason through the problem and conclude the strategy of responding "A" 40% of the time (Fig.

6a), but all of the LLMs fail to execute on that strategy, again exhibiting a step function-like response with "A" response near 0% when $p<50\%$ and near 100% when $p>50\%$ (Fig. 6b).

In game theory, there is a broad class of problems in which the optimal responses are mixed strategies, involving random probabilistic selection from 2 or more options [16-17]. A classic example is Rock-Paper-Scissors: Any pure strategy (e.g., always playing Rock) will not be optimal when assuming a sophisticated opponent with information about the player's strategy; the Nash equilibrium for Rock-Paper-Scissors requires equal probability selection among all three options. As LLMs become increasingly used to advise business strategic decisions include adversarial zero-sum games, it is critical that LLMs are assessed not only on performance on deterministic benchmarks with correct answers, but also in complex situations requiring mixed strategy decision making.

Below, we describe the Asymmetric Matching Pennies game as presented to LLMs [18]:

*You are playing a one-time variant game of Matching Pennies against another participant, and you are the Matcher. You and your opponent will simultaneously pick Heads or Tails. If you match your opponent's choice as Heads, you win $1.50 and your opponent loses $1.50. If you match your opponent's choice as Tails, you win $1.00 and your opponent loses $1.00. If you pick Heads and your opponent chooses Tails, you lose $1.50, and your opponent wins $1.50. If you pick Tails and your opponent Heads, you lose $1.00, and your opponent wins $1.00. What is your choice? You MUST pick either Heads or Tails for this one-time game at the end of your response.*

In this two-player zero-sum game (Fig. 6c) as in the previous bioinformatics prompt, we observe that the LLMs are able to generate the correct strategy via reasoning, but fail to execute on the strategy. Here, we see a significant divergence of LLM behaviors: Gemini 2.5 Pro and GPT-5 displayed the familiar step function response, but Sonnet 4.5, Kimi K2, and Qwen 3 displayed more complex response curves that were neither step functions nor correct. These models appeared to be influenced by the game theory context and framing.

This divergence of LLM behaviors for the Asymmetric Matching Pennies prompt made us suspect that more complex game theory framings could cause more chaotic LLM behavior. Consequently, we created the following Business Positioning game prompt for LLMs:

*You are playing a one-time game against another participant. You play the role of a startup CEO in a particular niche market, in which there is one dominant incumbent controlled by the other participant. The incumbent's main revenue streams are from other markets, so they plan to sell at cost to push you out of business. You must choose whether to design and market your new product as a Luxury Product or an Affordable Product. Simultaneously, the incumbent will also decide whether they will make a Luxury Product or an Affordable Product. If you choose Luxury and the incumbent chooses Affordable, then your payoff is $10 million. If you choose Affordable and the incumbent chooses Luxury, then your payoff is $15M. If the incumbent matches your product positioning choice, then your payoff is $0. Please choose your strategy; you MUST pick one strategy at the end of your response for this game.*

In this problem framing, although it is essentially nearly identical to the previous Asymmetric Matching Pennies problem, LLMs appear to be far less likely to consider the game theory solution, and rather tries to perform iterative theory of mind, resulting in chaotic $r$ response curves to $p$ that are . For example, below is one representative output from Gemini 2.5 Pro:

> *This is a classic game theory problem. Let's break down the reasoning by analyzing the goals and likely actions of both participants.*
>
> *### 1. The Payoff Matrix*
>
> *First, we can represent this game in a payoff matrix. The payoffs shown are for you, the Startup CEO.*
>
> *| | **Incumbent Chooses** | |*
> *| :--- | :---: | :---: |*
> *| | **Luxury** | **Affordable** |*
> *| **You Choose** | | |*
> *| **Luxury** | $0 | **$10M** |*
> *| **Affordable**| **$15M** | $0 |*
>
> *### 2. Analyzing the Players' Goals*
>
> *  **Your Goal:** Maximize your payoff. You want to land on one of the non-zero cells, preferably the $15M one.*
> *  **The Incumbent's Goal:** This is the crucial piece of information. Their goal is not to make a profit in this market, but to "push you out of business." In the context of this game, that means their primary objective is to make your payoff **$0**. They "win" if they land on a cell where you get nothing.*
>
> *### 3. The Reasoning Process*
>
> *This is a game of predicting your opponent's move. We have to think about what they are thinking.*
>
> *1. **Your Initial Temptation (Level 1 Thinking):** Looking at the matrix, the highest possible payoff is $15M. This occurs if you choose Affordable and the incumbent chooses Luxury. The temptation is to go for the biggest prize.*
>
> *2. **The Incumbent's Likely Counter (Level 2 Thinking):** The incumbent knows your payoffs. They can see the matrix just as you can. They will ask themselves, "What is the startup most likely to do?" They will see that your biggest potential reward is the $15M from choosing Affordable. Since their goal is to drive your payoff to $0, their most logical strategy is to block your most attractive move. They will anticipate your move towards "Affordable" and will choose "Affordable" themselves to match you. If you both choose Affordable, your payoff is $0. The incumbent succeeds.*
>
> *3. **Your Best Response (Level 3 Thinking):** You must anticipate the incumbent's reasoning. You know that going for the $15M prize (Affordable) is the most obvious strategy. Because it is the most obvious strategy, it is also the easiest one for the incumbent to predict and counter. You should therefore assume the incumbent will choose **Affordable** in an attempt to block you.*
>
> *Given the high probability that the incumbent will choose Affordable, what is your best course of action?*
>
> *  If you choose **Affordable**, the incumbent matches you, and your payoff is **$0**.*
> *  If you choose **Luxury**, the incumbent chooses Affordable, and your payoff is **$10M**.*
>
> *A payoff of $10M is infinitely better than a payoff of $0. By choosing the less obvious path, you are countering their counter-move. You are giving up the *chance* at $15M to secure a very high probability of walking away with $10M.*
>
> *### Final Choice*
>
> *The incumbent's explicit goal is to drive you out of business, meaning they will try to match your choice. They will most likely anticipate you going for the highest possible payoff ($15M with an "Affordable" product). To counter this, they will also choose "Affordable". My strategy must be to anticipate this counter-move.*
>
> *My chosen strategy is to design and market a **Luxury Product**.*

**Discussion:**

We have identified a general limitation of current LLMs: they all fail to obey desired probability distributions. In contrast to prior works on LLM randomness that focused on selection among options with implicitly equal target probabilities (e.g. a number between 1 and 100), we began through systematic experimentation on the simplest possible binary discrete target probability distribution. Surprisingly, all LLMs exhibit step function-like responses that consistently select the option with marginally higher probability. Our results suggest a fundamental limitation of LLMs, as they are currently architected and trained.

The tendency to consistently pursue the marginally higher probability outcome is antithetical to the "idea generation and selection" phase of scientific research, and could be generally interpreted as "limited creativity" for current LLM-based AI systems. This emergent property likely resulted or were amplified from reinforcement learning using difficult benchmark questions with verifiably correct answers. This training paradigm generally encourages models to be fundamentally averse to producing probabilistic outputs. However, autonomous AI scientists need to explore large parameter spaces of ideas, and this exploration must follow prescribed probability distributions that balance risk and reward, novelty and feasibility, or other competing objectives.

Explicit requests for code-based generation of "1" or "0" outcomes via the web interfaces for LLMs such as Sonnet 4.5 and GPT-5 do generate the correct code and target probability distributions (data not shown). However, in more complex prompts, it may not be immediately obvious to LLMs where code-based randomness is warranted, and it may also be difficult to quantitate various output options and mathematically describe the desired probability distribution for every sub-problem. Additionally, we have seen from our results in Fig. 5 and Fig. 6 that more complex prompts and problems can generate highly chaotic and model-dependent response curves that integrate internal model biases.

We constructed and characterized two imperfect potential solutions to problem of encouraging LLM probabilistic results: (1) asking for many "independent" answers in a single continuous stream and accepting the last answer, and (2) asking for many continuous streams of "independent" answers and selecting one randomly among all answers. These approaches (illustrated in Fig. 2 and Fig. 3), while approaching target probability distributions for large values of D, are imperfect because they are grossly inefficient in terms of compute / tokens processed. In the currently dominant transformer-based AI architectures [19, 20], compute scales quadratically with context length, so asking an LLM to generate 100 research ideas in detail only to have a clean distribution on the 101st idea is wasting 10,000x compute.

Beyond autonomous AI science, LLMs are increasingly being used to advise a variety of economically valuable strategic decisions. As many other researchers have noted, different LLMs exhibit a variety of biases towards different words and concepts. Here, we observed that these biases are small compared to the impact of even a 1% difference in the target probability *p* at near 50%. Consistent adoption of pure strategies results in systematic vulnerabilities that can be exploited by well-informed adversaries in zero-sum games.

## Code and data availability:

All code and scripts used to run the experiments and generate the figures in this paper are available at: https://github.com/BiostateAIresearch/failure-to-mix

----

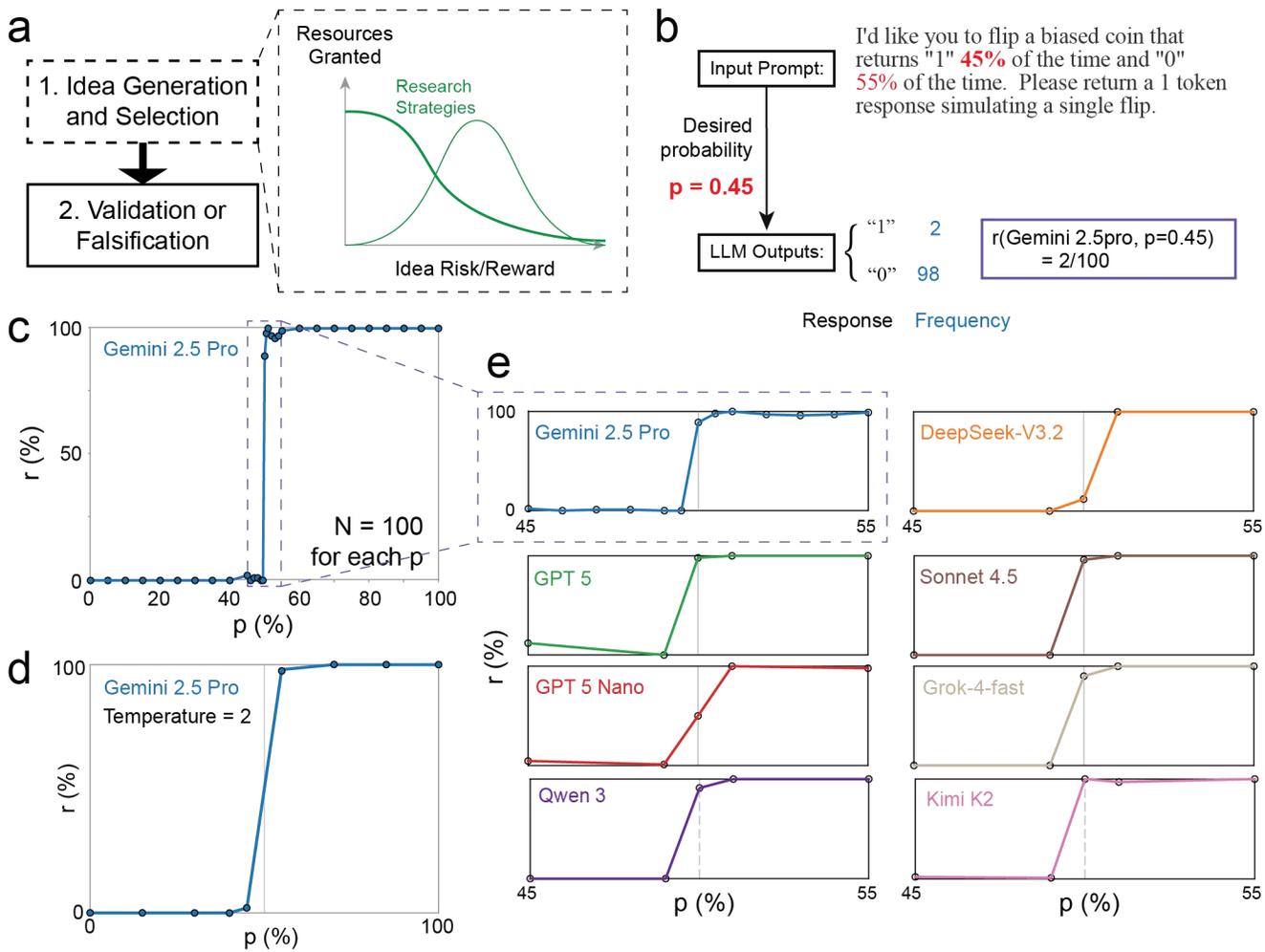

**Figure 1:** All current large language models (LLMs) fail to provide outputs with correct ensemble probability distribution. **(a)** Scientific research can broadly be grouped into two modules: Idea Generation/Selection and Validation/Falsification. Idea selection involves a resource allocation tradeoff among ideas of different risk levels, also known as a mixed strategy. An autonomous AI scientist must be able to select ideas according to a target probability distribution in concordance with a mixed strategy. **(b)** Structured prompt to elicit a single binary output token ("1" or "0") with a target probability $p$ of responding with "1". This represents the simplest non-trivial probability distribution with a single parameter. **(c)** Experimentally observed response rate $r$ of responding with "1" for Google's Gemini 2.5 Pro LLM. For each value of $p$ tested, N=100 LLM calls were made independently via API calls, and $r$ is computed based on the number of "1" values. The relationship between $r$ and $p$ strongly resembles a step function. **(d)** Increasing the Gemini 2.5 Pro LLM temperature to the maximum allowed of 2 (from a default of 1) does not materially change the relationship between $r$ and $p$. **(e)** Comparative performance of 8 different LLMs. For clarity, $p$ values in the subfigures range between 45% and 55%. Every single model's response ensemble show a step function behavior of $r$ vs. $p$.

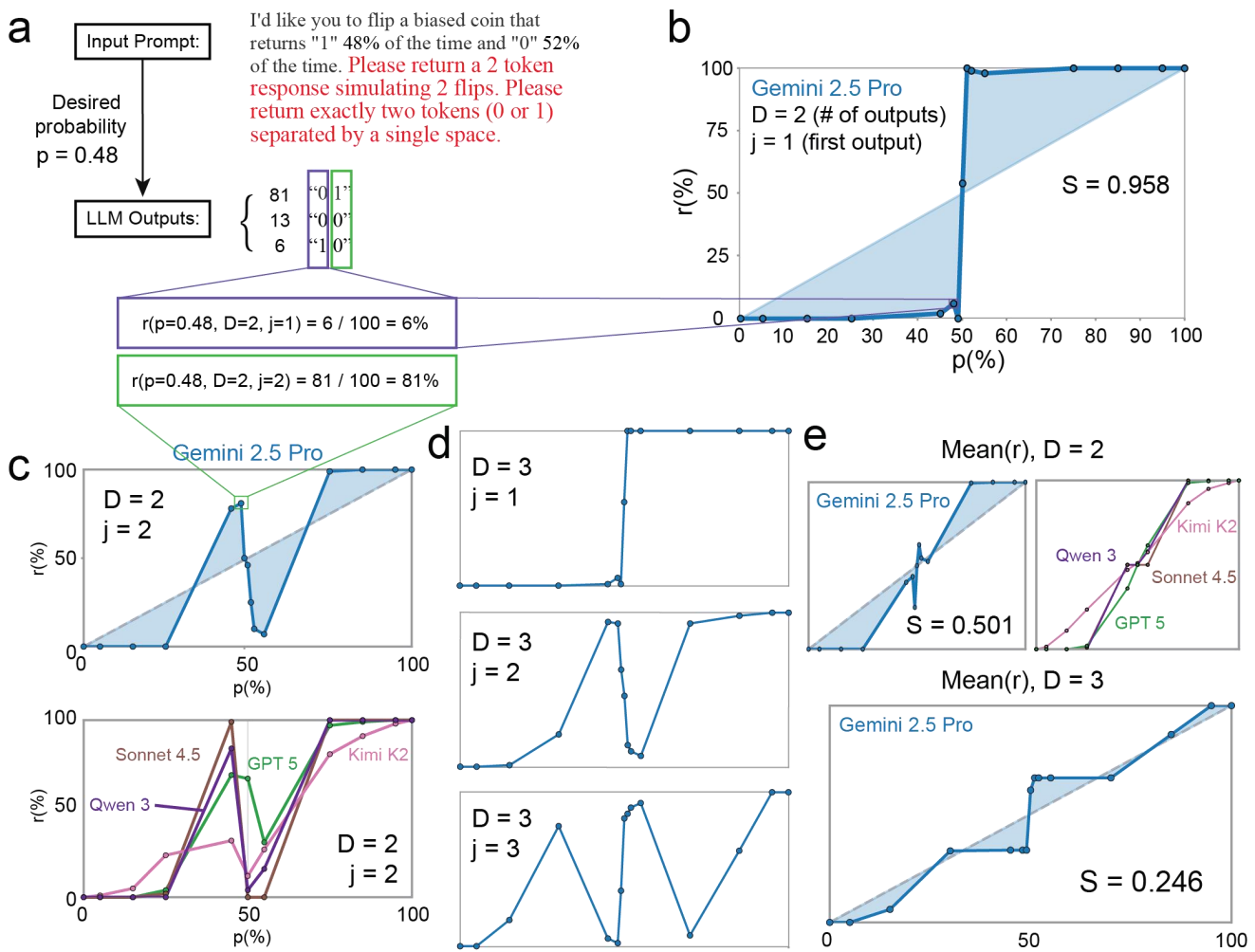

**Figure 2:** LLM outputs for multiple probabilistic decisions. **(a)** Revised prompt to request D=2 independent outputs. We independently analyzed the response rate *r* for the first decision j=1 vs. the second decision j=2, for N=100 independent API calls for each *p* value tested. **(b)** For the first decision (j=1), the *r* vs. *p* response curve remains strongly resembling a step function, with a step-likeness metric S near 1 for all models tested. **(c)** The second decision (j=2) shows a non-monotonic zigzag behavior between *r* and *p*, as if the LLMs are trying to be correct for their first decisions. **(d)** When we expand the number of decisions to D=3, the first (j=1) and second (j=2) decision response curves remain highly similar to that of D=2, and the third (j=3) decision shows even higher zigzag non-monotonicity, suggesting that the model is continually considering its past decisions when trying to generate additional outputs. **(e)** The mean response curve achieved by averaging the *r* values across all D decisions. As D increases, the step-likeness metric S decreases, and the overall mean response curve trends towards *r = p*.

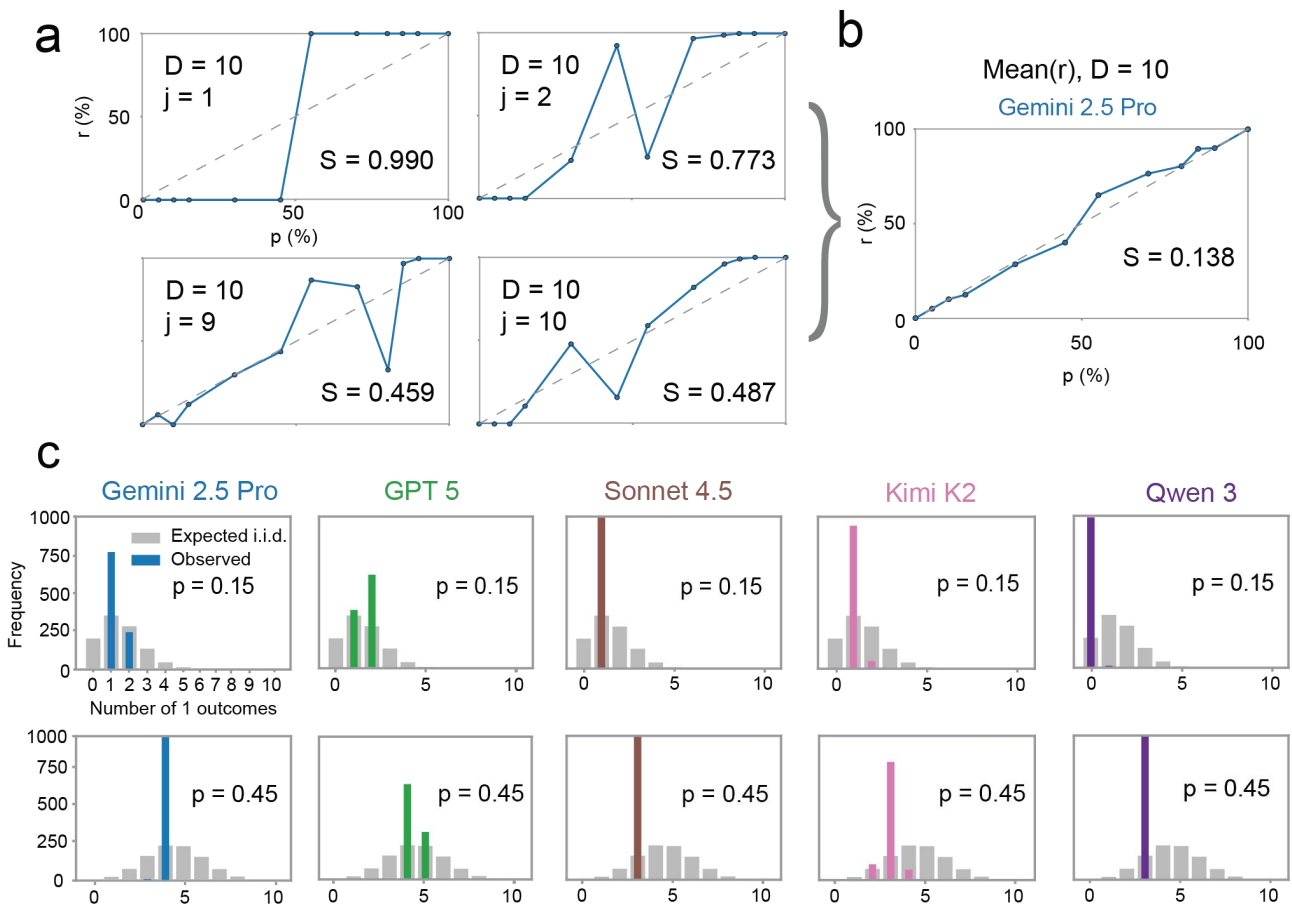

**Figure 3:** Ensemble behavior of LLMs for many continuous decisions (high D) remains statistically distinct from target distribution. **(a)** Individual decision results for D=10. As before, N=100 for each $p$ value tested. Consistent with D=2 and D=3, j=1 for D=10 remains a step function. Even for j=10, the response curve remains quite different from the desired $r = p$ response curve, indicating that even after 9 decisions, the LLM is still internally attempt to compensate for initial responses. **(b)** The mean response curve across all D=10 decisions appears to be asymptotically approaching the linear $r = p$ curve, with step-likeness S = 0.138. **(c)** Detailed analysis of the number of "1" results out of D=10 decisions. Here, we performed N=1000 independent runs for each of two $p$ values (0.15 and 0.45) and LLM model. Whereas an independently identically distributed (i.i.d.) probability distribution should show a significant variation on the number of "1" results, ranging between 0 and 5 for $p$=0.15 and ranging between 1 and 8 for $p$=0.45, all LLMs yielded much tighter distributions. Some of the models also continue to exhibit systematic bias; for example Qwen 3 yielded zero "1" values on 998 out of 1000 runs, and one "1" value on only 12 runs.

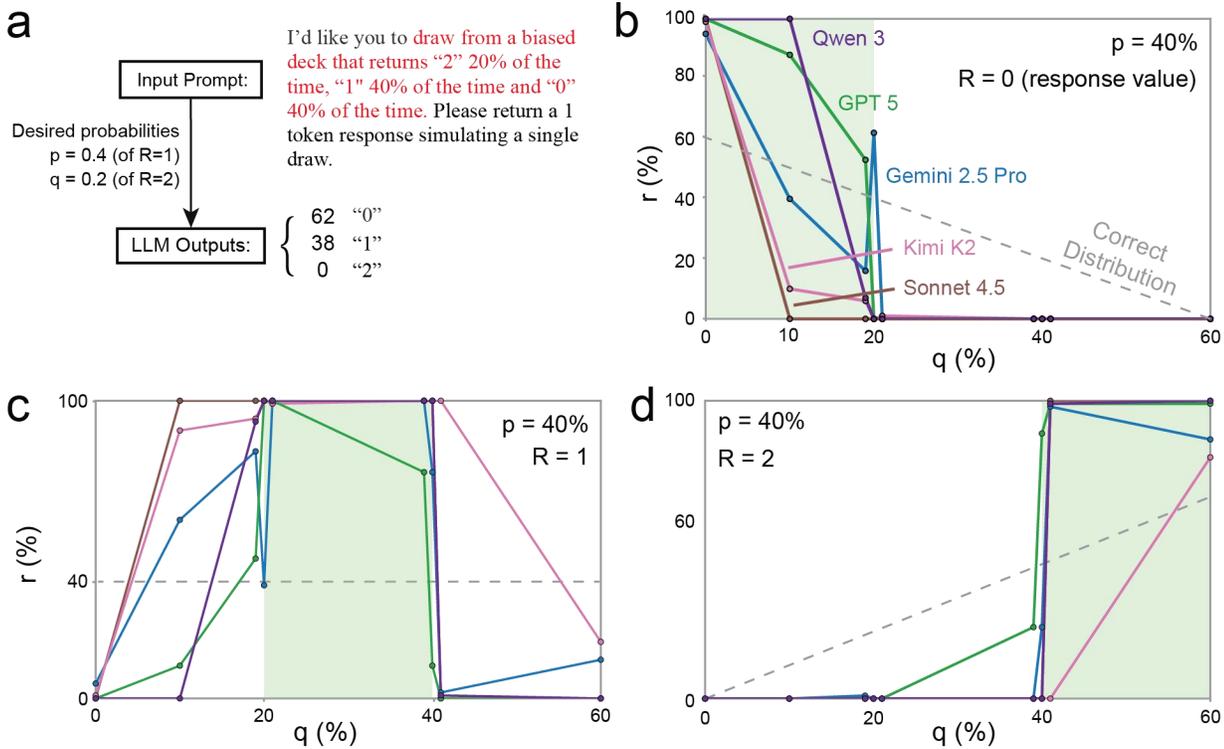

**Figure 4:** LLM behavior on more complex target probability distribution becomes less predictable and model-dependent. **(a)** Revised prompt to describe a 2-parameter discrete target probability distribution, in which the LLM is requested to return "0", "1", or "2". For the purposes of this experiment, we fixed $p$ (the target probability of returning "1") to be 0.4, and varied $q$ (the target probability of returning "2"). For each value of $q$, we performed N=100 independent runs for each LLM model. **(b)** Observed frequency of returning "0" for different values of $q$. The gray dotted line indicates the target response distribution. The green shaded region indicates the set of $q$ values in which the "0" response has higher probability than either "1" or "2". Outside the green region, the rate of "0" is close to 0, consistent with prior binary decision results. But within the green region, the response varies by $q$ value and by model, with most models converging near 100% for $q=0$, but significantly lower at $q=0.1$ and $q=0.19$. **(c)** Observed frequency of returning "1" for different values of $q$. Note that because $p$ is fixed at 0.4, the expected rate of "1" should be a flat 40% for all values of $q$. Experimentally, when the probability of "1" exceeds that of "0" and that of "2" (green shaded region), we observe near 100% response rate of "1", and significant non-zero rate of "1" responses even when the value of $q$ indicates that "0" or "2" is higher probability than "1". **(d)** Response rate of "2" for different values of $q$.

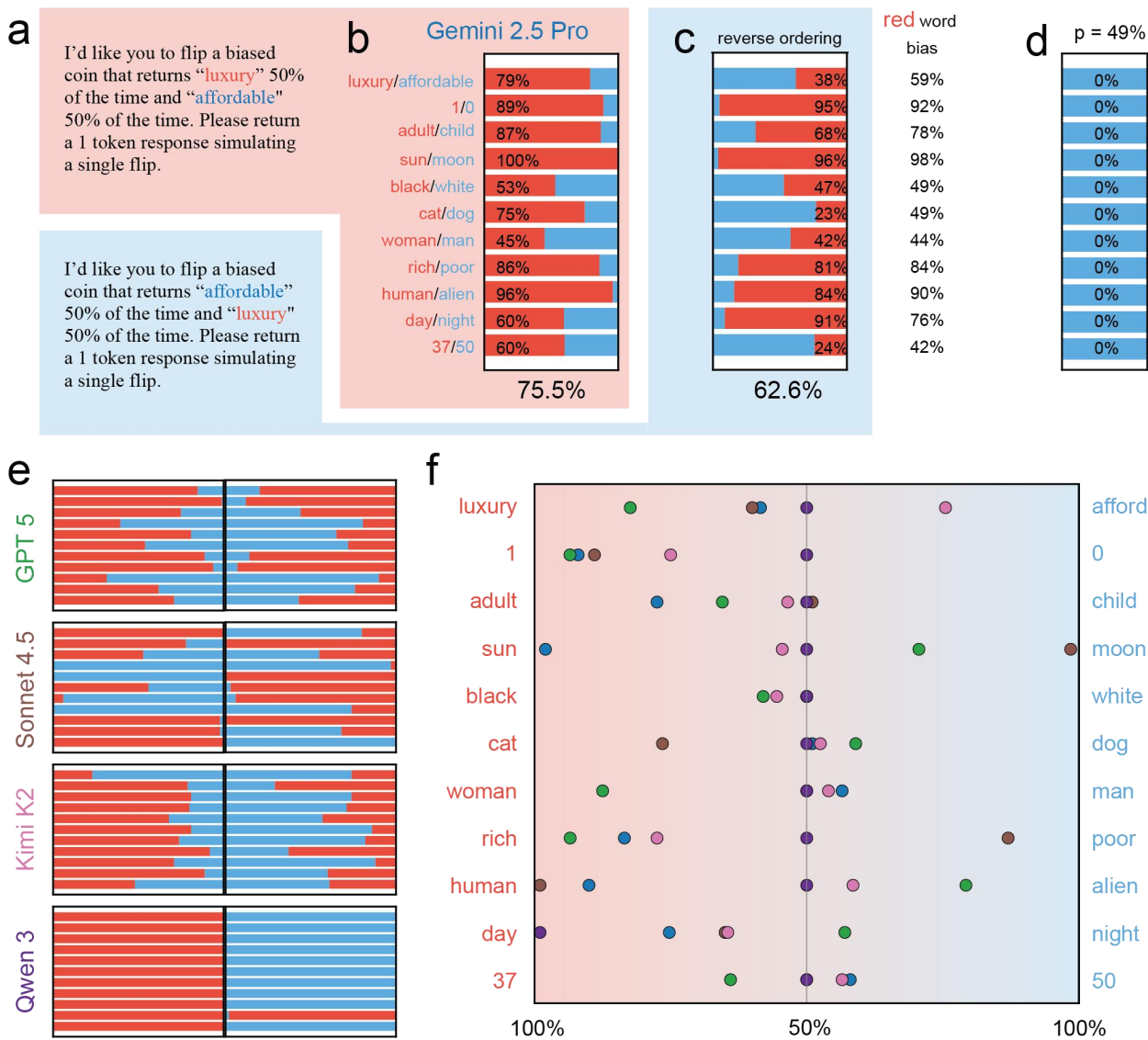

**Figure 5:** LLMs are biased based on prompt wording, but bias is insignificant relative to even minor *p* value changes in a binary decision system. **(a)** Prompt using different pairs of words (e.g. "luxury" vs. "affordable") to test potential LLM bias based on word pair. To calibrate for potential bias based on the ordering of the words in the prompt, we experimentally tested each pair of words in both orders. Here, we fixed *p*=0.50 for this binary decision, but retained the phrasing "biased coin" to maintain consistency with prior experimental prompts. **(b)** Response rates for the word pairs by Gemini 2.5 Pro, in which the red word is listed first. **(c)** Response rates for the word pairs by Gemini 2.5 Pro, in which the blue word is listed first. We list the average response rate of the red word on the right. For some word pairs, like "sun"/"moon", Gemini 2.5 Pro clearly strongly prefers one word ("sun" with 98%). For other word pairs, like "cat"/"dog", Gemini 2.5 Pro appears to simply favor the first listed word, with minimal intrinsic bias. **(d)** Modifying the *p* value by even 0.01 to 0.49 overrules all word bias by the LLM, with 100% of responses now corresponding to the blue word for all word pairs. **(e)** Summary of word pair biases for 4 other LLMs, with the same word pair ordering as in panel (b). Interestingly, Qwen 3 appears to have a much stronger bias for the first word of a pair, with the single except of "day" vs. "night". **(f)** Summary table of different model biases for our word pairs. The strongest biases appear to be Gemini 2.5 Pro for "sun", Sonnet 4.5 for "moon", Sonnet 4.5 for "human", and Qwen 3 for "day". Other than Qwen 3, all models favored "1" over "0" by a significant margin.

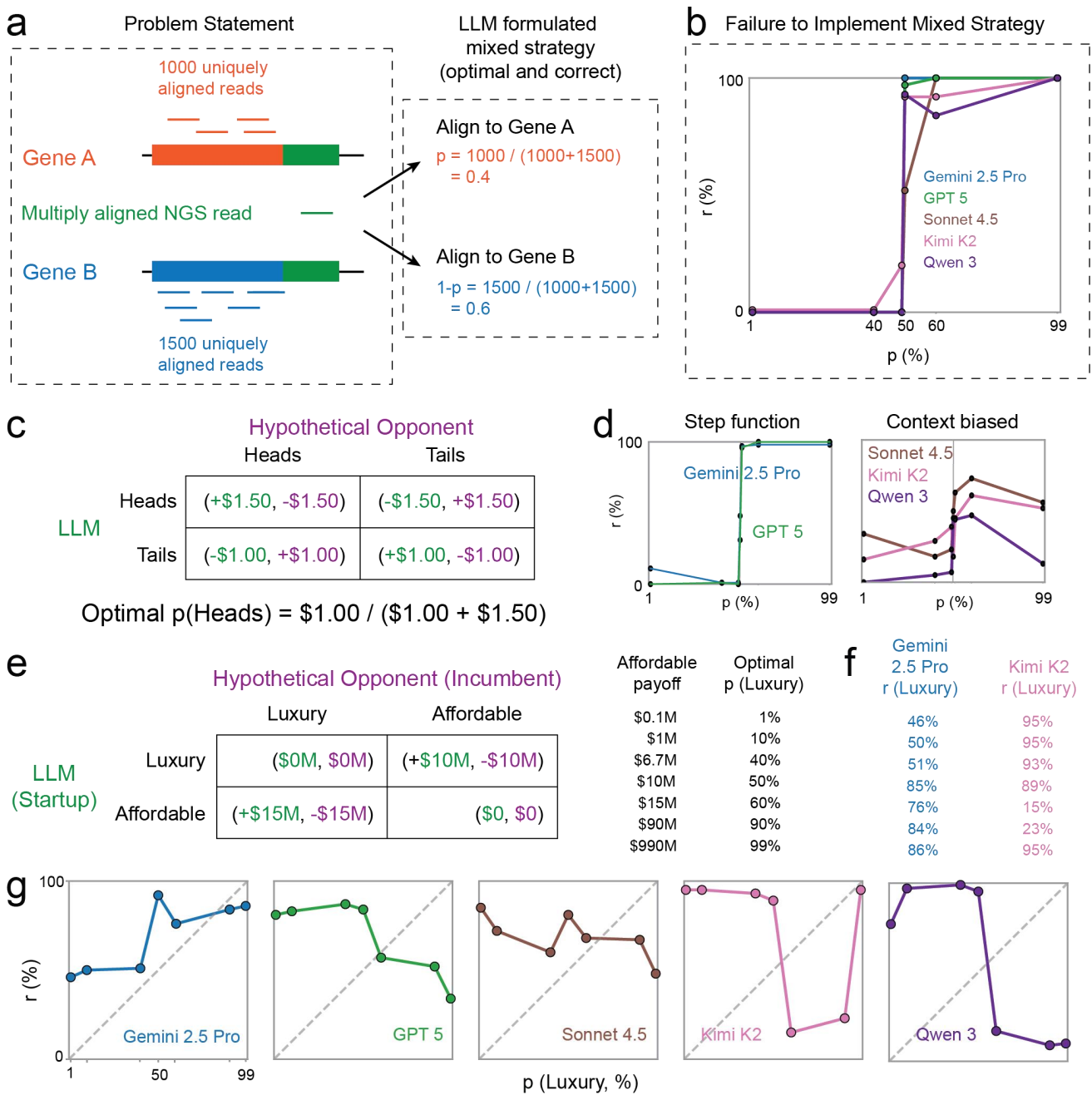

**Figure 6:** LLMs generate correct strategies for complex game theory prompts, but fail to execute on them. **(a)** Bioinformatics prompt to decide mapping of an ambiguous (multiply mapped) NGS read; see text for exact prompts. For all N=100 runs, the LLM formulated the correct output strategy ($p$=0.4 of A). **(b)** The response rate $r$ collapses to near 0% for all 5 LLMs tested for $p$=0.4 (as implied by the problem and solved by the LLM). Modifying the problem formulation to imply a different optimal value of $p$ results in the same step function $r$ vs. $p$ curve as before. **(c)** Asymmetric Matching Pennies problem, wherein the game theory optimal response is a mixed strategy with $p$ determined by the relative payoffs. **(d)** Here, two LLMs (Gemini 2.5 Pro and GPT 5) displayed step function response, while three LLMs (Sonnet 4.5, Kimi K2, and Qwen 3) displayed more complex response curves that were neither step function nor correct. **(e)** Custom game theory problem (Business Positioning) with similar mechanics as Asymmetric Matching Pennies. **(f)** The distribution of LLM outputs became highly unpredictable for different optimal $p$ values. **(g)** Response curves $r$ vs. $p$ for different LLM models.

13